%% file: main.tex
\documentclass[letterpaper, 10 pt, conference]{ieeeconf}  

\IEEEoverridecommandlockouts                              

\usepackage{graphics} 
\usepackage{epsfig} 
\usepackage{mathptmx} 
\usepackage{times} 
\usepackage{amsmath} 
\usepackage{amssymb}  
\usepackage{todonotes}
\usepackage{etoolbox}
\usepackage[font=small,labelfont=bf]{caption} 
\usepackage{hyperref}
\usepackage{sidecap}

\usepackage{booktabs}
\usepackage{pifont}
\newcommand{\cmark}{\ding{51}}%
\newcommand{\xmark}{\ding{55}}%
\usepackage{graphicx}
\usepackage{subfig}
\usepackage{xspace}
\definecolor{mygreen}{RGB}{15, 157, 88}
\definecolor{myred}{RGB}{219, 68, 55}

\usepackage{hyperref}
\let\svthefootnote\thefootnote
\newcommand\freefootnote[1]{%
  \let\thefootnote\relax%
  \footnotetext{#1}%
  \let\thefootnote\svthefootnote%
}

\newcommand{\ourshort}{USEEK\xspace}

\title{\LARGE \bf
\ourshort: Unsupervised SE(3)-Equivariant 3D Keypoints for \\
Generalizable Manipulation
}

\author{Zhengrong Xue$^{1,3}$, Zhecheng Yuan$^{2,1}$, Jiashun Wang$^{4}$, Xueqian Wang$^{2}$, Yang Gao$^{2,5,1}$, Huazhe Xu$^{2,5,1}$ 
}

\begin{document}

\twocolumn[{%
\renewcommand\twocolumn[1][]{#1}%
\maketitle
\input{figure/teaser.tex}
}]



\thispagestyle{empty}
\pagestyle{plain}

\freefootnote{$^1$Shanghai Qi Zhi Institute. $^2$Tsinghua University. $^3$Shanghai Jiao Tong University. $^4$Carnegie Mellon University. $^5$Shanghai AI Lab.}
\freefootnote{Contact:\tt\small{ xuezhengrong@sjtu.edu.cn, huazhe\_xu@mail.tsinghua.edu.cn}.}

\input{text/0_abstract.tex}
\input{text/1_intro.tex}
\input{text/2_related.tex}
\input{text/4_method.tex}
\input{text/5_exp_vis.tex}
\input{text/6_exp_mani.tex}
\input{text/7_conclusion.tex}

{\small
\bibliographystyle{abbrv}
\bibliography{main.bib}
}

\end{document}

%% file: figure/teaser.tex
\begin{center}
\centering
\captionsetup{type=figure}
\includegraphics[width=\linewidth]{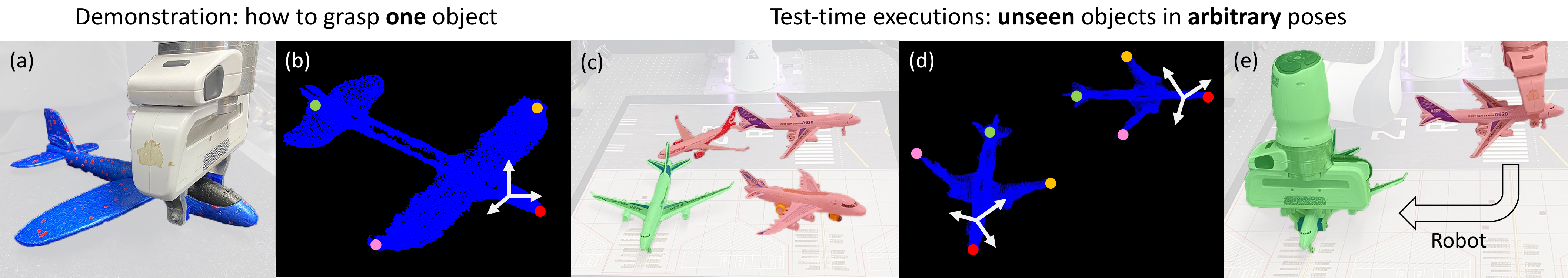}
\captionof{figure}{(a) Given an object point cloud and the mere demonstration of a functional grasping pose, (b) \ourshort infers a set of keypoints and the task-relevant local coordinate frame. (c) When tested, unseen objects within the category in \textcolor{myred}{initial poses} unobserved at the training time are shown to the robot. The robot is required to pick and then place it to an arbitrary \textcolor{mygreen}{target pose}. (d) The properties of intra-category alignment and SE(3)-equivariance make \ourshort generalizable to novel shapes and poses. (e) With the help of keypoints and local coordinate frames, the robot manages to transfer the functional knowledge and execute the manipulation tasks.}

\label{fig:teaser}
\end{center}

%% file: text/0_abstract.tex
\begin{abstract}
     Can a robot manipulate intra-category unseen objects in arbitrary poses with the help of a mere demonstration of grasping pose on a single object instance? In this paper, we try to address this intriguing challenge by using \ourshort, an \underline{u}nsupervised \underline{SE}(3)-\underline{e}quivariant \underline{k}eypoints method that enjoys alignment across instances in a category, to perform generalizable manipulation. \ourshort follows a teacher-student structure to decouple the unsupervised keypoint discovery and SE(3)-equivariant keypoint detection. With \ourshort in hand, the robot can infer the category-level task-relevant object frames in an efficient and explainable manner, enabling manipulation of any intra-category objects from and to any poses. Through extensive experiments, we demonstrate that the keypoints produced by \ourshort possess rich semantics, thus successfully transferring the functional knowledge from the demonstration object to the novel ones. Compared with other object representations for manipulation, \ourshort is more adaptive in the face of large intra-category shape variance, more robust with limited demonstrations, and more efficient at inference time. Project website: \url{https://sites.google.com/view/useek/}.
    
\end{abstract}

%% file: text/1_intro.tex
\section{Introduction}
\label{sec:intro}

When three-year old children think of an object, they recognize it not only as the object itself but also as a symbol for the category~\cite{deloache1987rapid}. The innate talents of humans to generalize, according to developmental psychology, are known as symbolic functioning~\cite{mccune1981toward}. In the context of robotics, the same desire to generalize begs the research question: does there exist a control method that achieves generalizable manipulation across object poses and instances?

With the access to the full 3D geometry of the object, the pipeline for robotic manipulation has long been mature --- template matching~\cite{fischler1981random,besl1992method,rusinkiewicz2001efficient} for perception while trajectory optimization~\cite{brooks1983planning,ratliff2009chomp} and inverse kinematics~\cite{chiaverini1994review} for execution. However, these manipulation skills often suffer from intra-category shape variance as common hand-crafted techniques for template matching may fail to generalize.

 To enable  intra-category any-pose manipulation, an object representation that achieves category-level generalization is crucial. Existing representations can be roughly classified into three kinds: 6-DOF pose estimators~\cite{wang2019normalized,wang2019densefusion,wang20206,wen2022you,li2021leveraging}, 3D keypoints~\cite{suwajanakorn2018discovery,shi2021skeleton,manuelli2019kpam,manuelli2021keypoints,chen2021unsupervised}, and dense correspondence models~\cite{schmidt2016self,florence2018dense,sundaresan2020learning,simeonov2022neural}.
Despite the disparities in form, their ultimate goals are consistent --- to determine the local coordinate frame of the object. Thus, we tend to view those representations as different abstraction levels of an object. Among them, 6-DOF pose estimators provide the highest level of abstraction by predicting the object frame directly. However, they are often regarded as not generalizable enough for large shape variance in robotic manipulation tasks~\cite{manuelli2019kpam,simeonov2022neural}.
Recently, dense correspondence models implicitly define an object by approximating a continuous function that maps either 2D pixels or 3D points to spatial descriptors. While these spatial descriptors preserve abundant geometric details, the object frames directly guiding manipulation cannot be acquired gratis from the dense correspondence representations.

Compared with the aforementioned representations, 3D keypoints enjoy the benefits of both practicality and simplicity: its semantic correspondences are more informative than 6-DOF poses; its succinct expression is more efficient than dense correspondence models.
Despite the advantages of keypoints, for the task of intra-category any-pose robotic manipulation, we may further require the keypoints to possess the following properties:

\begin{itemize}
    \item (i) \textit{Anti-occlusion.} The keypoints should be repeatable in the face of self-occlusion. Thus, we prefer raw 3D inputs (i.e., point clouds) to multi-view images.
    \item (ii) \textit{Unsupervised.}
    The keypoints should be obtained in an unsupervised or self-supervised manner to avoid the costs and biases from human annotations.
    \item (iii) \textit{Aligned across instances.}
    The semantic correspondence of keypoints across instances within a certain category is essential for category-level generalizable manipulation.
    \item (iv) \textit{SE(3)-equivariant.} The keypoints are further desired to be equivariant w.r.t. the translations and rotations of the objects in the 3D space because the objects in the wild can appear in any poses.
\end{itemize}

In this paper, we propose a framework that utilizes 3D keypoints for intra-category any-pose robotic manipulation. At the heart of this framework are the discovered 3D keypoints that boast all of the four desired properties. Specifically, we propose a novel teacher-student architecture for \underline{u}nsupervised \underline{SE}(3)-\underline{e}quivariant \underline{k}eypoint~(USEEK) discovery. We then first evaluate \ourshort against state-of-the-art keypoint discovery baselines through visual metrics. Next, we leverage \ourshort to enable a robot to pick up an intra-category object from a randomly initialized pose and then place it in a specified pose via one-shot imitation learning.
Despite the difficulty, rich semantics of the keypoints given by \ourshort enables the robot to execute pick-and-place by transferring the functional knowledge from limited demonstration to unseen instances in any poses. Quantitative and qualitative results in the simulator as well as on the real robot indicate that \ourshort is competent to serve as an object representation for generalizable manipulation tasks.

%% file: text/2_related.tex
\section{Related Work}
\label{sec:related}

\subsection{Object Representations for Manipulation}
\noindent\textbf{Explicit 6-DOF pose estimation.}
Pose estimation techniques start from the early works such as RANSAC~\cite{fischler1981random} or Iterative Closest Point (ICP)~\cite{rusinkiewicz2001efficient}. Though very efficient, these works usually struggle with the shape variance of unknown objects. Learning-based 6-DOF pose estimators~\cite{sahin2018category,wang2019normalized,wang20206,li2021leveraging} manage to represent an object on a category level. But when applied to robotic manipulation, they are often viewed as either ambiguous under large intra-category shape variance~\cite{manuelli2019kpam}, or incapable to provide enough geometric information for control~\cite{simeonov2022neural}.

\noindent\textbf{Dense correspondence.}
In contrast to the explicit pose prediction, dense correspondence methods~\cite{schmidt2016self,florence2018dense,florence2019self,sundaresan2020learning,simeonov2022neural} define an object in a continuous and implicit way. One example is the recently proposed Neural Descriptor Fields~(NDF)~\cite{simeonov2022neural}, which encodes the spatial relations of external rigid bodies and the demonstrated object. Effective as it is for few-shot imitation learning, NDF is inefficient because it has to regress the descriptor fields of hundreds of query points via iterative optimization~\cite{kingma2014adam}.

\noindent\textbf{3D keypoints.} The use of 3D keypoints for control is extensively studied in computer vision~\cite{suwajanakorn2018discovery,li2019usip,you2022ukpgan,shi2021skeleton}, robotics~\cite{manuelli2021keypoints,manuelli2019kpam,gao2021kpam}, and reinforcement learning~\cite{vecerik2021s3k,chen2021unsupervised}. However, we find that none of the existing methods shown in Table~\ref{tab:kp-compare} meets all the requirements we have listed that are beneficial to the task of generalizable robotic manipulation.


\input{table/kp_compare.tex}

\subsection{SE(3)-Invariant/Equivariant Neural Networks}
The concepts of SE(3)-invariant and SE(3)-equivariant are sometimes intertwined.
For the function of keypoint detection, we require it to be \textit{SE(3)-invariant} if the function selects the indices of points in the point cloud; otherwise, we require it to be \textit{SE(3)-equivariant} if the function returns the coordinates of the keypoints.
In this paper, we take the second interpretation, requiring the 3D keypoints to be \textit{SE(3)-equivariant}. Noting that we can easily handle translations by normalizing the center of mass of the point cloud to a specified original point, the challenges are mainly entangled with rotations, i.e., SO(3)-invariance/equivariance.

PRIN/SPRIN~\cite{you2020pointwise,you2021prin} and Vector Neurons~\cite{deng2021vector} are recently proposed SO(3)-invariant networks that directly takes point clouds as inputs. PRIN extracts rotation invariant features by absorbing the advantages of both Spherical CNN~\cite{cohen2018spherical} and PointNet~\cite{qi2017pointnet}-like networks. SPRIN improves PRIN in sparsity and achieves state-of-the-art performance. Concurrently, Vector Neurons enjoy SO(3)-equivariance by extending neurons from 1D scalars to 3D vectors and providing the corresponding SO(3)-equivariant neural operations.

%% file: table/kp_compare.tex
\begin{table}
\centering
\begin{tabular}{lcccc}
    \toprule
    \multicolumn{1}{r}{Property} & (i) & (ii) & (iii) & (iv)\\
    \midrule
    KeypointNet~\cite{suwajanakorn2018discovery} & \xmark & \cmark & \cmark & \xmark \\
    USIP~\cite{li2019usip} & \cmark & \cmark & \xmark & \xmark \\
    UKPGAN~\cite{you2022ukpgan} & \cmark & \cmark & \xmark & \cmark \\
    Skeleton Merger~\cite{shi2021skeleton} & \cmark & \cmark & \cmark & \xmark \\
    \midrule
    kPAM~\cite{manuelli2019kpam} and its variants & \cmark & \xmark & \cmark & \xmark \\
    Keypoints into the Future~\cite{manuelli2021keypoints} & \xmark & \cmark & \cmark & \xmark \\
    \midrule
    S3K~\cite{vecerik2021s3k} & \cmark & \cmark & \cmark & \xmark \\
    Keypoint3D~\cite{chen2021unsupervised} & \cmark & \cmark & \xmark & \xmark \\
    \bottomrule
\end{tabular}
\caption{We compare the features of recently proposed 3D keypoint detectors in the field of computer vision, robotics, and reinforcement learning.}
\vspace{-0.2cm}
\label{tab:kp-compare}
\end{table}

%% file: text/4_method.tex
\section{Methods}
\label{sec:method}

\input{figure/pipeline.tex}

We present a framework that utilizes the \underline{u}nsupervised, \underline{S}\underline{E}(3)-\underline{e}quivariant \underline{k}eypoints~(\ourshort) for intra-category any-pose object manipulation. The keypoints from \ourshort are first used to specify the task-relevant local coordinate frame of a demonstration object. Then, they generalize to the corresponding points from objects of unseen shapes within the same category and in unobserved poses at training time. Leveraging such keypoints, we manage to transfer the task-relevant frames, and finally perform motion planning algorithms to manipulate the objects.

\subsection{Preliminaries on Keypoints}

We first define a keypoint detector as $f(\cdot)$ that maps an object point cloud $\mathbf{P}$ to an ordered set of keypoints $\mathbf{p}$:
\begin{align}
    f(\mathbf{P}):\mathbb{R}^{N\times3}\to\mathbb{R}^{K\times3},
\end{align}
where $N$ is the number of points and $K$ is the number of keypoints. The function is SE(3)-equivariant if for any point cloud $\mathbf{P}$ and any rigid body transformation $(\mathbf{R},\mathbf{t})\in\mathrm{SE(3)}$, the following equation holds:
\begin{align}
    f(\mathbf{R}\mathbf{P}+\mathbf{t}) \equiv \mathbf{R} f(\mathbf{P})+\mathbf{t}.
\end{align}
Furthermore, keypoints detected are considered as category-level if they can best represent the shared geometric features of a category of objects. 

\subsection{\ourshort: a Teacher-Student Framework}
\label{sec:ts}
To develop a keypoint detector that is both category-level and SE(3)-equivariant, we propose \ourshort that has a teacher-student structure. 
The teacher network is a category-level keypoint detector that can be pre-trained in a self-supervised manner. The student network consists of an SE(3)-invariant backbone. 

Conceptually, the major merit of the teacher-student networks is to decouple the learning process, where each network is only responsible for the property that it is most adept at. Moreover, the SE(3)-invariant networks are usually harder to train. Hence, the teacher-student structure might alleviate the burden of the student network in the process of keypoint discovery.
These are the central reasons why the teacher-student structure is essential and why the simpler approaches (shown in Section~\ref{sec:vis-setup}) cannot achieve competitive results. Next, with the blueprint for \ourshort established, we instantiate it with concrete details.

\noindent\textbf{The teacher network.}
In the teacher network, every keypoint is considered as the weighted sum of all the point coordinates in the cloud.
To produce the desired weight matrix $\mathbf{W}\in\mathbb{R}^{K\times N}$, we extract the PointNet++~\cite{qi2017pointnet++} encoder from Skeleton Merger~\cite{shi2021skeleton}, a state-of-the-art category-level keypoint detector.
The multiplication of the weight matrix and the input cloud directly gives the predicted keypoints
\begin{align}
    \mathbf{p}=\mathbf{W}\mathbf{P}.
\end{align}
We follow the same self-supervised training procedure as shown in~\cite{shi2021skeleton} to pre-train the PointNet++ module.

The predicted keypoints by the teacher network are used to generate pseudo labels for the student network.
Since nearby points share the same semantics, all the points within a distance of a certain keypoint are considered as candidates for the corresponding keypoint and therefore marked with positive labels; the rest are the negative ones. With altogether $K$ keypoints detected in an $N$-point cloud, the final pseudo label can be regarded as a $K$-channel binary segmentation mask $\mathbf{M}\in\mathbb{R}^{K\times N}$.

\noindent\textbf{The student network.} The student network of \ourshort utilizes SPRIN~\cite{you2021prin}, a state-of-the-art SE(3)-invariant backbone, to produce SE(3)-equivariant keypoints.
The SPRIN module takes as input the canonical object point cloud and predicts the labels generated by the teacher network. 

For training, the student network optimizes a Binary Cross Entropy~(BCE) loss between the per-point $K$-channel binary predictions and the corresponding pseudo labels. 
To deal with imbalanced negative labels over positive ones, we perform importance sampling~\cite{tokdar2010importance}. Besides, we highlight that the whole training procedure does not require any SE(3) data augmentations because the SE(3)-invariant backbone could automatically generalize to unseen poses. 

When tested, the predictions are post-processed to produce the final keypoints. We take the \texttt{argmax} operation, which means the point that has the highest confidence value is selected as the detected keypoint for each of the $K$ segmentation classes. Moreover, we take Non-Maximum Suppression~\cite{neubeck2006efficient} to encourage sparse locality of the keypoints.

\subsection{From Keypoints to Task-Relevant Object Frames}

The essence of generalizable manipulation is arguably to transfer the functional knowledge from known object(s) to the unknowns.
In this section, we exhibit an easy-to-execute yet effective procedure that leverages the detected keypoints to determine the task-relevant object frames.

Taking the airplane in Figure~\ref{fig:frame} (Left) as an illustrative example, the four detected keypoints lie on the nose, the tail, the left wingtip, and the right wingtip, respectively.
With clear semantics, we can easily set up simple rules so as to establish the frame: the $x$-axis is parallel to the connected line of the two wingtips; the $z$-axis is parallel to the connected line of the nose and the tail; the $y$-axis is vertical to both the $x$-axis and the $z$-axis; the origin is the projection of the demonstrated grasping position on the $z$-axis.
Thanks to the intra-category alignment property of \ourshort, once the rules on \textit{one} specific object are set up, they instantly adapt to \textit{all} the other instances.
In this sense, the average labor spent per instance is negligible.
In Figure~\ref{fig:frame} (Right), we provide the category of the guitar as an additional example.

Notably, unlike previous works~\cite{manuelli2019kpam,simeonov2022neural}, \ourshort avoids any searching or optimization process when inferring object frames, thus dramatically reducing the computational costs.

\input{figure/object_frame.tex}

\subsection{One-Shot Imitation Learning with \ourshort}
Equipped with keypoints and task-relevant object frames, we are now ready for category-level manipulation. The task is to pick up an unseen object from a randomly initialized SE(3) pose and place it to another specified pose. To showcase the full potential of \ourshort, we follow a more challenging setting of \textit{one}-shot imitation learning rather than the \textit{few}-shot setting commonly seen in prior works~\cite{simeonov2022neural}, since a decreased number of demonstrations calls for increased robustness and consistency of the proposed object representation.

Specifically, the demonstration $\mathcal{D}=(\mathbf{P}_\mathrm{demo},\mathbf{X}_\mathrm{demo}^G)$ is merely the point cloud of an object $\mathbf{P}_\mathrm{demo}$ and a functional grasping pose of the \underline{G}ripper $\mathbf{X}_\mathrm{demo}^G$ in the form of a coordinate frame. Given the demonstration $\mathcal{D}$, \ourshort infers the task-relevant coordinate frame of the demonstration \underline{O}bject $\mathbf{X}_\mathrm{demo}^O$. Then, we calculate the rigid body transformation from the \underline{O}bject to the \underline{G}ripper ${}^O\mathbf{T}_\mathrm{demo}^G$, s.t.
\begin{align}
    \mathbf{X}_\mathrm{demo}^G=\mathbf{X}_\mathrm{demo}^O {}^O\mathbf{T}_\mathrm{demo}^G,
\end{align}
where all the poses and transformations are in the homogeneous coordinates. 
Assuming the object is rigid and the grasp is tight, ${}^O\mathbf{T}_\mathrm{demo}^G$ is general for the category. Thus, we rewrite it as ${}^O\mathbf{T}^G$ for simplicity.

At test time, the observation $\mathcal{O}=(\mathbf{P}_\mathrm{init},\mathbf{P}_\mathrm{targ})$ consists of the point cloud of an unseen object in an arbitrary initial pose $\mathbf{P}_\mathrm{init}$ and another point cloud indicating the target pose $\mathbf{P}_\mathrm{targ}$. Note that it is not required that the object in the target pose is the same as the one in the initial pose. \ourshort infers the object frame in the initial pose $\mathbf{X}_\mathrm{init}^O$ and the frame in the target pose $\mathbf{X}_\mathrm{targ}^O$. Then, with ${}^O\mathbf{T}^G$ prepared, we can easily calculate the poses of the gripper for pick and place:
\begin{align}
    \mathbf{X}_\mathrm{pick}^G=\mathbf{X}_\mathrm{init}^O{}^O\mathbf{T}^G,\\
    \mathbf{X}_\mathrm{place}^G=\mathbf{X}_\mathrm{targ}^O{}^O\mathbf{T}^G.
\end{align}
Finally, we leverage off-the-shelf motion planning~\cite{kuffner2000rrt} and inverse kinematics~\cite{saif_sidhik_2020_4320612} tools to execute the predicted poses.

%% file: figure/pipeline.tex
\begin{figure*}[t]
\centering
    \includegraphics[width=0.98\textwidth]{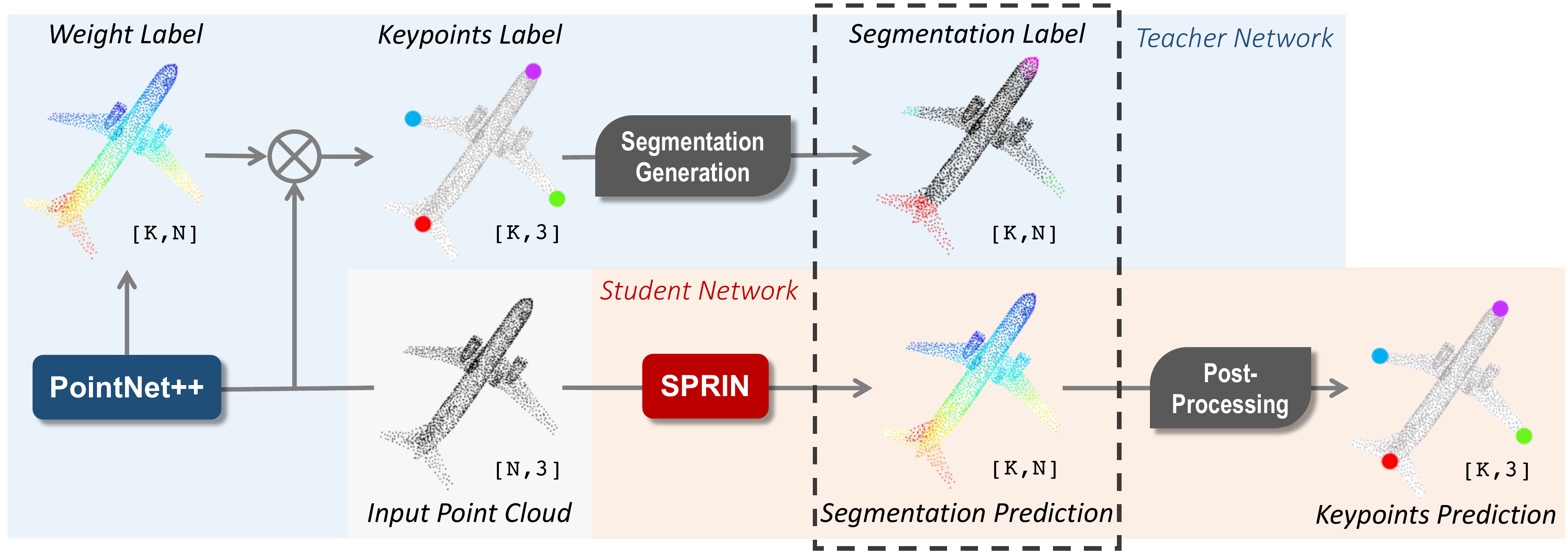}
    \caption{The pipelines of \ourshort, which follows a teacher-student architecture. All the ``labels'' are pseudo ground-truth labels generated by the teacher network, free from any additional human annotations. The PointNet++~\cite{qi2017pointnet++} module is with fixed parameters, extracted from a pre-trained Skeleton Merger~\cite{shi2021skeleton}. The SPRIN~\cite{you2021prin} network is to be optimized in the training process. Binary Cross Entropy (BCE) loss is used for loss computation.}
    \vspace{-0.2cm}
    \label{fig:pipeline}
\end{figure*}

%% file: figure/object_frame.tex
\begin{figure}[t]
\centering
    \includegraphics[width=0.45\textwidth]{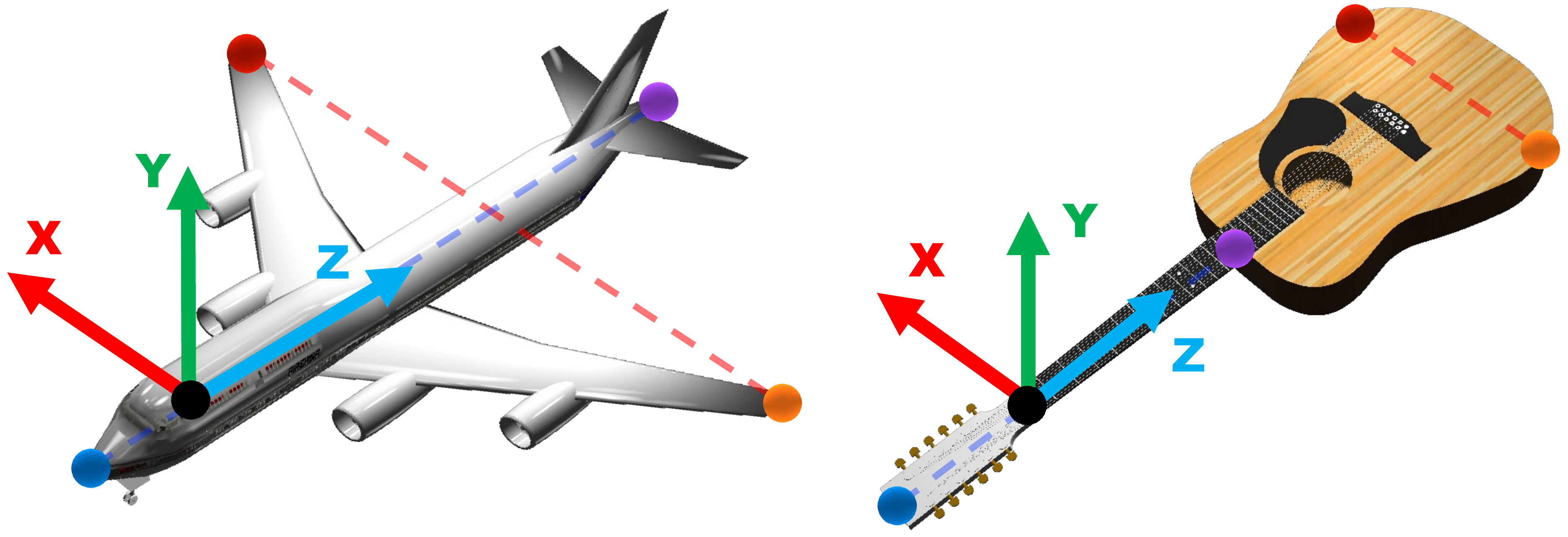}
    \caption{The category-level task-relevant coordinate frames for manipulation can be acquired from the keypoints detected by \ourshort together with few human decided priors on only one object.}
    \label{fig:frame}
    \vspace{-0.2cm}
\end{figure}

%% file: text/5_exp_vis.tex
\section{Experiments: Semantics of Keypoints}
\label{sec:exp-vis}

\input{figure/qualitative.tex}

In this section, we evaluate whether the keypoints detected by \ourshort are with proper and accurate semantics when the input point clouds are under SE(3) transformations.


\input{table/vis_exp.tex}

\subsection{Setup and Baselines}
\label{sec:vis-setup}

The experiments are conducted on the KeypointNet~\cite{you2020keypointnet} dataset, where keypoints with category-level semantic labels are annotated by experts. We use the mean Intersection over Unions (mIoU)~\cite{teran20143d} score to measure the alignment between the predictions and the human annotations. To evaluate the property of SE(3)-equivariance, the inputs and their annotations are under the same random SE(3) transformations.
We compare \ourshort with the following methods:
\begin{itemize}
    \item \textit{Intrinsic Shape Signatures (ISS).} ISS~\cite{zhong2009intrinsic} is a classic hand-crafted 3D keypoint detector. 
    \item \textit{Skeleton Merger.} The Skeleton Merger is trained on ShapeNet~\cite{chang2015shapenet} with canonical point clouds. 
    \item \textit{Skeleton Merger w/ data augmentation.} We apply SE(3) data augmentations to the training dataset. 
    \item \textit{Skeleton Merger w/ ICP.} During test time, we randomly take one instance in canonical pose on the training dataset as the template, and adopt ICP~\cite{rusinkiewicz2001efficient} initialized with RANSAC~\cite{fischler1981random} for point cloud registration.
    \item \textit{Skeleton Merger w/ SPRIN encoder.} The encoder of Skeleton Merger is replaced with an SE(3)-invariant SPRIN encoder. 
    \item \textit{\ourshort w/ KL divergence.} The SE(3)-invariant backbone in \ourshort is slightly revised to predict weight matrices. It is optimized via the Kullback–Leibler (KL) divergence~\cite{kullback1951information} between the predicted weights and the pseudo weight labels.
\end{itemize}

Additionally, we evaluate the \textit{Teacher Network} of \ourshort on the \textit{canonical} KeypointNet dataset w/o SE(3) transformations. This auxiliary configuration reflects the quality of the semantics that \ourshort could learn from.

\subsection{Results and Discussion}
\label{sec:vis-result}

The qualitative results of keypoints detected by \ourshort are shown in Figure~\ref{fig:qualitative}. Under SE(3) transformations and with large shape variance, the keypoints are well aligned across the category and identify semantic parts that are akin to human intuition. The quantitative results given in Table~\ref{tab:vis-exp} reveal that \ourshort substantially outperforms all the other baselines by a large margin. 
Surprisingly, \ourshort approaches Teacher Network performance on the categories of airplanes and guitars, and even surpasses it on knives.
In fact, it is reasonable that the student in \ourshort may excel its teacher, given the elaborately designed label generation mechanism where surrounding points of a predicted keypoint are all considered as the keypoint candidates.

For the other baselines, it is as expected that baselines such as Skeleton Merger w/ data augmentation are unable to produce semantically meaningful keypoints because it is extremely hard, if not impossible, for the naive PointNet++ encoder to capture invariant patterns from arbitrary SE(3) transformations.
Further, we also notice that stronger baseline methods such as Skeleton Merger w/ SPRIN encoder is not as capable as \ourshort. 
We attribute its failure to the fact that the SE(3)-invariant guarantee of the encoder often results in more parameters in the network as well as much more difficulty in training. Compared with the baselines, \ourshort enables the training process by following a teacher-student structure, which transforms unsupervised causal discovery of latent keypoints to a simpler supervised learning task.

%% file: figure/qualitative.tex
\begin{figure*}[t]
\centering
    \includegraphics[width=\textwidth]{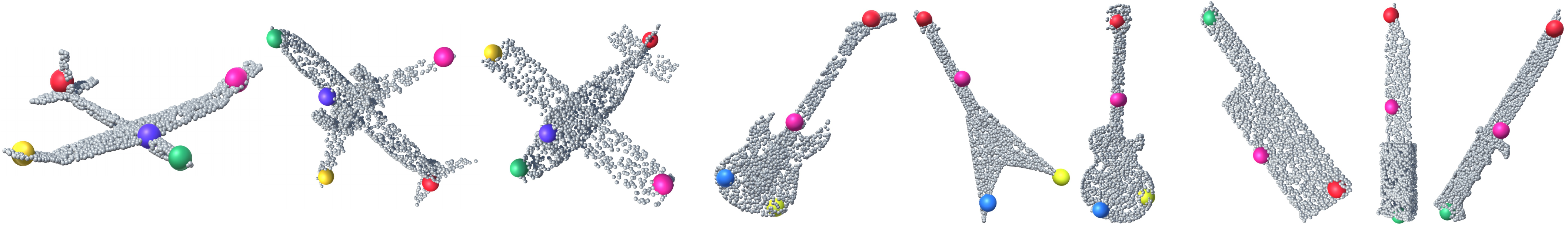}
    \caption{Qualitative results of the keypoints detected by \ourshort. The input point clouds are under random SE(3) transformations. The color of the keypoints stands for the predicted category-level semantic correspondence~(i.e., keypoints of a category are color-aligned).}
    \vspace{-0.2cm}
    \label{fig:qualitative}
\end{figure*}

%% file: table/vis_exp.tex
\begin{table}
\centering
\begin{tabular}{lcccc}
    \toprule
     & Airplanes & Chairs & Guitars & Knives\\
    \midrule
    ISS~\cite{zhong2009intrinsic} & 31.9 & 9.9 & 34.6 & 35.8\\
    \midrule
    Skeleton Merger~\cite{shi2021skeleton} & 19.9 & 10.7 & 24.2 & 18.8\\
    \quad w/ data augmentation & 14.0 & 7.3 & 10.9 & 21.2\\
    \quad w/ ICP~\cite{rusinkiewicz2001efficient} & 22.2 & 10.0 & 25.8 & 17.2\\
    \quad w/ SPRIN~\cite{you2021prin} encoder & 74.7 & 18.4 & 41.1 & 45.6\\
    \midrule
    \ourshort w/ KL divergence & 55.7 & 40.1 & 64.9 & 17.0\\
    
    \midrule
    \ourshort (ours) & \textbf{85.5} & \textbf{53.8} & \textbf{70.2} & \textbf{60.3}\\
    \midrule
    \textcolor{gray}{Teacher Network} & \textcolor{gray}{87.0} & \textcolor{gray}{65.3} & \textcolor{gray}{71.1} & \textcolor{gray}{40.3}\\
    \bottomrule
\end{tabular}
\caption{mIoU scores of keypoints detected by \ourshort and the baselines on the SE(3) KeypointNet~\cite{you2020keypointnet} dataset. The best results are shown in \textbf{bold} type. In addition, \textit{Teacher Network} tested on the canonical dataset is also included but marked in \textcolor{gray}{gray} for distinction.}
\vspace{-0.2cm}
\label{tab:vis-exp}
\end{table}


%% file: text/6_exp_mani.tex
\section{Experiments: Keypoints for Manipulation}
\label{sec:exp-mani}

\input{figure/manipulation.tex}

In this section, we evaluate the power of \ourshort as an object representation for generalizable manipulation. We conduct experiments in both simulated and real-world environments where \ourshort is utilized to perform category-level pick-and-place via one-shot imitation learning.

\subsection{Setups}
We build simulated environment mimicking the physical setup in PyBullet~\cite{coumans2019}. For the real-world environment, we use a Franka Panda robot arm for manipulation and four Intel RealSense D435i depth cameras at each corner of the table for the capture of point clouds. We use wooden wedges, metal holders, or plasticine-kneaded holders to support the objects so that they can be placed in arbitrary poses. The real-world experiment setup is visualized in Figure~\ref{fig:manipulation}.

The robot needs to pick up unseen objects in randomly initialized SE(3) poses and place them in a specified target pose. For achieving this task, only one demonstration of a functional grasping pose on a single object is provided for each category. For greater diversity, we design the tasks on three categories: 1) for airplanes, we demonstrate two distinct functional poses (i.e., in the front and in the rear) to inspect whether the robot can acts accordingly with the demonstration; 2) for guitars, the robot is asked to transfer the guitar from one metal stand to the other by holding its neck. This task is more challenging because the neck of a guitar is delicate and thus asks for very precise control;
3) for knives, the robot needs to pick up a knife on the table and use it to chop tofu. The chopping direction is inferred from a given target pose. 

\subsection{Baselines and Evaluation Metrics}
To show the effectiveness of \ourshort for manipulation, we perform comparison in the simulated environment with ICP~\cite{rusinkiewicz2001efficient} initialized with RANSAC~\cite{fischler1981random} coarse registration and Neural Descriptor Fields (NDF)~\cite{simeonov2022neural}. ICP estimates the frame of an unseen object by trying to align it to the given demonstration object. NDF is a recently proposed state-of-the-art dense correspondence representation for category-level manipulation. NDF encodes the relation of every SE(3) pose of external rigid bodies and a demonstration object through a set of query points. The task-relevant coordinate frame of an unseen object is determined by regressing its descriptor field to the demonstrated one.

We measure the success rates for both grasping itself and the whole process of pick-and-place. Specifically, the grasping is successful if the gripper stably takes the object off the table with a vertical clearance of at least 10 centimeters. For pick-and-place, the transnational and rotational tolerance of the final pose is 5 centimeters and 0.2 radian respectively for each degree of freedom.
The experiments are performed on 100 different object instances randomly taken from ShapeNet~\cite{chang2015shapenet} without cherry-picking for each category. Besides, we also present the average inference time that each method takes to perform one execution.

\subsection{Training Details}
We train both \ourshort and the NDF baseline on the ShapeNet~\cite{chang2015shapenet} dataset and directly deploy the models to the manipulation tasks. 
We use Adam~\cite{kingma2014adam} for optimization with a default learning rate of 0.001.
While NDF could handle multiple categories with one unified model, we find the performance of the given pre-trained model drops significantly because of the large shape variance in the chosen categories. Therefore, we train independent models of each category for both \ourshort and NDF for fair comparison.

\subsection{Simulation Experiments}
\input{table/manipulate.tex}

The success rates of ICP~\cite{rusinkiewicz2001efficient}, NDF~\cite{simeonov2022neural}, and \ourshort are shown in Table~\ref{tab:mani}. \ourshort significantly outperforms the two baselines in the challenging one-shot imitation learning setting, showcasing its strengths as an object representation for generalizable manipulation. In comparison, ICP as a template matching method is virtually unable to deal with large intra-category shape variance and cannot provide accurate information for manipulation. Meanwhile, NDF manages to offer relatively decent results on the category of knives. But it is overall far less competitive than \ourshort. We observe a failure mode where NDF cannot find the correct grasping orientation. We attribute this to the difficulty in regressing the neural descriptors of a large number of ($\sim$500 in default) query points with only one demonstration available. Furthermore, for NDF, it takes 3.65 seconds on average to infer for one execution on an NVIDIA RTX 3070 GPU while for \ourshort, it only takes 0.11 seconds, indicating a more than 30$\times$ efficiency boost.

\input{figure/real_world_all_objects.tex}

\subsection{Real World Execution}
Finally, we validate \ourshort can be successfully deployed for generalizable manipulation on the real robot without sim-to-real finetuning.
We apply the RRT-Connect~\cite{kuffner2000rrt} algorithm for motion planning and the default methods from the \texttt{Panda Robot} library~\cite{saif_sidhik_2020_4320612} for executing inverse kinematics.

Quantitatively, a total of 50 executions are conducted on a series of novel airplanes of various appearances and materials, and in arbitrary initial poses. Among them, 37 trials end up with success, resulting in a success rate of 74\%. We find that \ourshort is robust to the quality degradation of raw point clouds taken from depth cameras. The results also show that \ourshort can well adapt to the color changing and shape variance in real-world scenes. A common failure mode is that \ourshort is unable to handle the severe artifact in the point clouds (e.g., half of the wing is missing) largely due to high reflective rates of the oil paint on the surface of some airplanes. We believe this can be alleviated in the future with the access to more advanced depth cameras.
Qualitatively, we present example execution trajectories for airplanes, guitars, and knives in Figure~\ref{fig:traj}. More results can be found in the supplementary video.

%% file: figure/manipulation.tex
\begin{figure*}[t]
\centering
    \includegraphics[width=\textwidth]{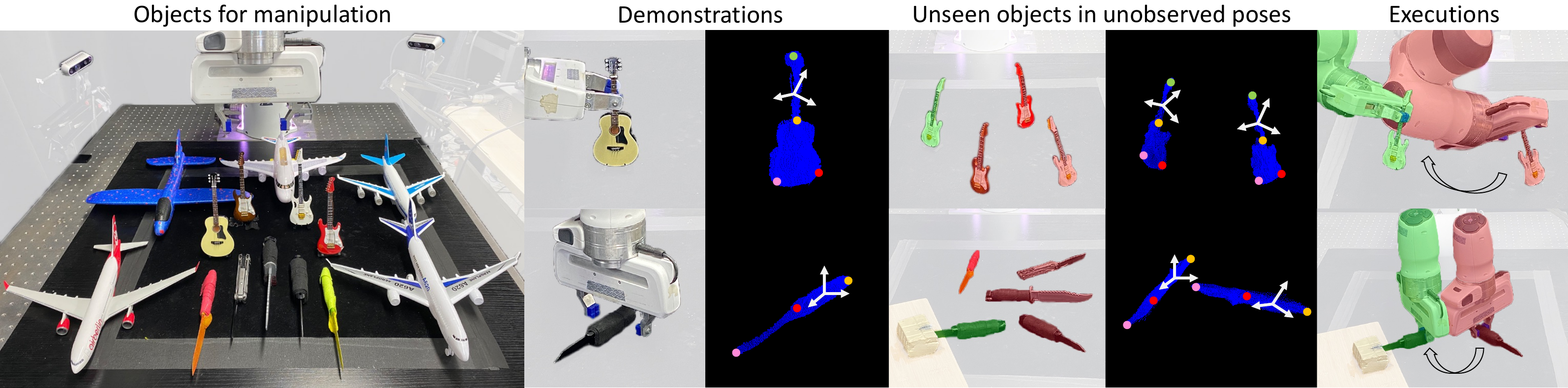}
    \caption{In real-world experiments, objects of large intra-category shape variance are manipulated. The alignment and SE(3)-equivariance properties enable \ourshort to transfer the functional knowledge from one demonstration object to various unseen objects in arbitrary poses.}
    \vspace{-0.2cm}
    \label{fig:manipulation}
\end{figure*}

%% file: table/manipulate.tex
\begin{table}
\centering
\begin{tabular}{@{}lcccccc@{}}
    \toprule
      & \multicolumn{2}{c}{Airplanes} & \multicolumn{2}{c}{Guitars} & \multicolumn{2}{c}{Knives}\\
     \cmidrule(lr){2-3} \cmidrule(lr){4-5} \cmidrule(lr){6-7} 
      &  Grasp & Overall &  Grasp & Overall &  Grasp & Overall\\
     \midrule
      ICP~\cite{rusinkiewicz2001efficient} & 0.08 & 0.00 & 0.00 & 0.00 & 0.01 & 0.00\\
      NDF~\cite{simeonov2022neural} & 0.25 & 0.04 & 0.22 & 0.01 & 0.71 & 0.61\\
      \ourshort & \textbf{0.90} & \textbf{0.81} & \textbf{0.89} & \textbf{0.69} & \textbf{0.93} & \textbf{0.85}\\
    \bottomrule
    
\end{tabular}
\caption{The success rates for grasping and the overall process of pick-and-place with the setting of one-shot imitation learning.}
\vspace{-0.2cm}
\label{tab:mani}
\end{table}

%% file: figure/real_world_all_objects.tex
\begin{figure}[t]
\centering
    \includegraphics[width=1.0\linewidth]{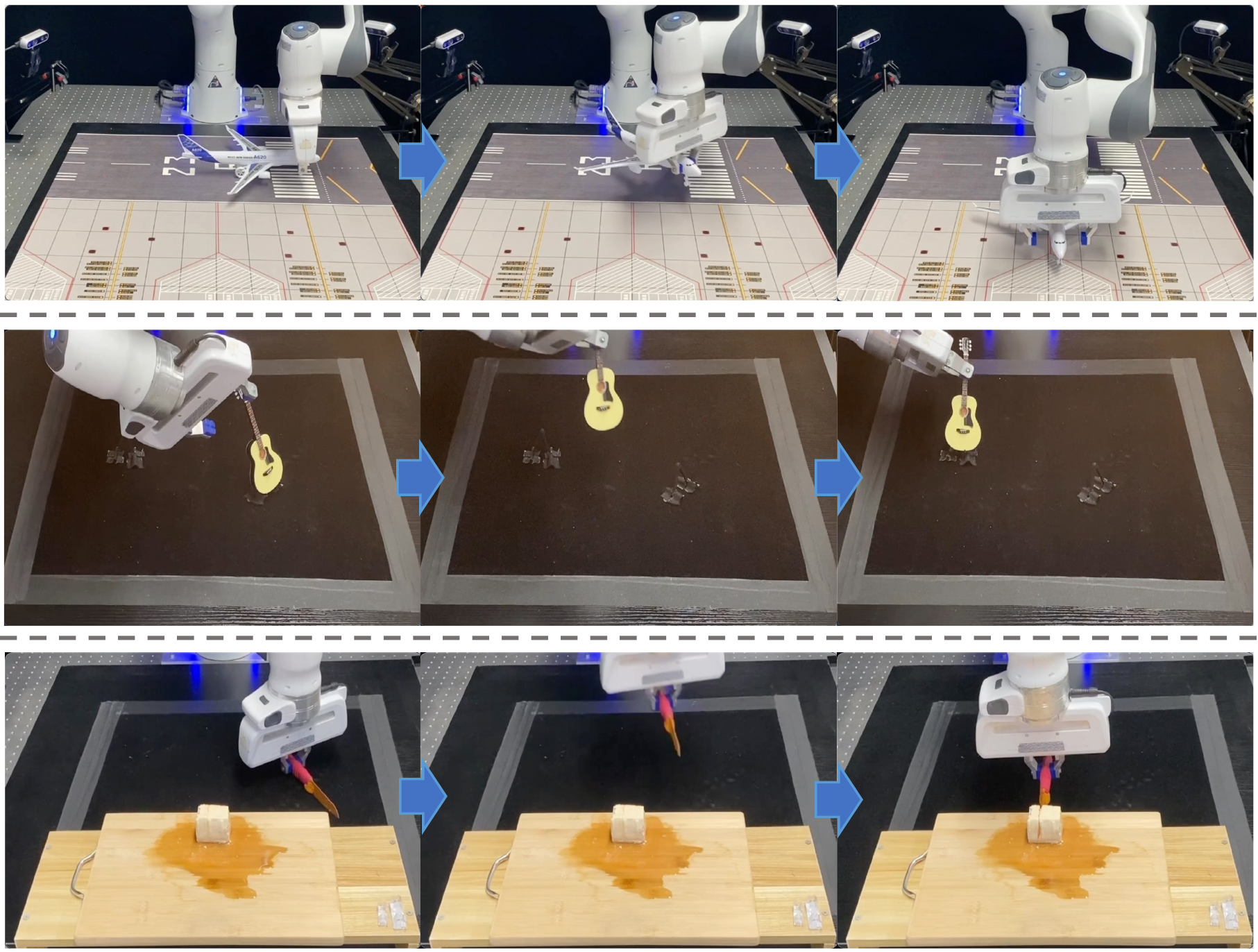}
    \caption{Leveraging \ourshort, we present the example execution trajectories of pick-of-place by manipulating the front of an airplane, the neck of a guitar, and the handle of a knife. }
    \vspace{-0.2cm}
    \label{fig:traj}
\end{figure}

%% file: text/7_conclusion.tex
\section{Conclusion}
\label{sec:conclusion}

We present \ourshort, an \underline{u}nsupervised \underline{S}\underline{E}(3)-\underline{e}quivariant \underline{k}eypoints detector to empower generalizable pick-and-place via one-shot imitation learning. \ourshort uses teacher-student structure so that keypoints with the desired properties can be acquired in an unsupervised manner. Extensive experiments show that the keypoints detected by \ourshort enjoy rich semantics, which enables the transfer of functional knowledge from one demonstration object to unseen objects with large intra-category shape variance in arbitrary poses. Further, \ourshort is found to be robust with limited demonstrations and efficient at inference time. Given all these advantages, we believe \ourshort is a capable object representation and has the potential to empower many manipulation tasks.